\documentclass[11pt]{article}

\usepackage[final]{acl}

\usepackage{times}
\usepackage{latexsym}

\usepackage[T1]{fontenc}

\usepackage[utf8]{inputenc}

\usepackage{microtype}

\usepackage{inconsolata}
\usepackage{xcolor}         
\usepackage{amsmath}
\usepackage{graphicx}
\usepackage{listings}
\usepackage{amsfonts}
\usepackage{subcaption}
\usepackage{multirow}
\usepackage{colortbl}
\usepackage{tabularx}
\usepackage{booktabs}
\usepackage{float}
\usepackage{wrapfig}
\usepackage{graphicx}
\usepackage{bm}
\usepackage{caption}
\usepackage{xurl} 
\definecolor{mypurple}{RGB}{253,245,250}
\usepackage{pifont}
\usepackage{enumitem}
\usepackage{verbatim}
\usepackage{fontawesome5}

\definecolor{pinkA}{RGB}{255,182,193}   
\definecolor{pinkB}{RGB}{255,105,180}   
\definecolor{blueA}{RGB}{173,216,230}   
\definecolor{blueB}{RGB}{135,206,235}   
\definecolor{figblue}{HTML}{ACDDE4}
\definecolor{figgray}{HTML}{7F7F7F}

\usepackage{enumitem}

\newcommand{\rewardmodel}{Agent-RRM}

\usepackage[most]{tcolorbox}

\newtcolorbox{promptbox}[1]{
    colback=gray!5, 
    colframe=black!75, 
    fonttitle=\bfseries,
    title=#1,
    arc=2pt,
    left=5pt,
    right=5pt,
    top=5pt,
    bottom=5pt,
    boxrule=0.5pt,
    width=\textwidth,
    enlarge bottom by=0.2cm
}

\title{Exploring Reasoning Reward Model for Agents}

\author{
 \textbf{Kaixuan Fan\textsuperscript{1,2}}\quad\!\!\!\!
 \textbf{Kaituo Feng\textsuperscript{1,2}}\quad\!\!\!\!
 \textbf{Manyuan Zhang\textsuperscript{2}\thanks{Project Leader.}}\quad\!\!\!\!
 \textbf{Tianshuo Peng\textsuperscript{1}}\quad\!\!\!\!
 \textbf{Zhixun Li\textsuperscript{3}}
 \\
 \textbf{Yilei Jiang\textsuperscript{1,2}}\quad\!\!\!\!
 \textbf{Shawn Chen\textsuperscript{2}}\quad\!\!\!\!
 \textbf{Peng Pei\textsuperscript{2}}\quad\!\!\!\!
 \textbf{Xunliang Cai\textsuperscript{2}}\quad\!\!\!\!
 \textbf{Xiangyu Yue\textsuperscript{1}\thanks{Corresponding Author.}}
\\
 \textsuperscript{1}MMLab, CUHK\quad\!\!
 \textsuperscript{2}Meituan\quad\!\!
 \textsuperscript{3}SEEM, CUHK
\\
\faGithub \textbf{Repository:} \url{https://github.com/kxfan2002/Reagent}
}

\begin{document}
\maketitle

\tcbset{
  agentbox/.style={
    colback=white,
    colframe=black!40,
    boxrule=0.5pt,
    arc=1.5mm
  },
  toolcallbox/.style={
    colback=blue!3,
    colframe=blue!50,
    boxrule=0.5pt,
    arc=1.5mm
  },
  tooloutputbox/.style={
    colback=blue!1,
    colframe=blue!30,
    boxrule=0.5pt,
    arc=1.5mm
  }
}

 \begin{abstract}

Agentic Reinforcement Learning (Agentic RL) has achieved notable success in enabling agents to perform complex reasoning and tool use.
However, most methods still relies on sparse outcome-based reward for training.
Such feedback fails to differentiate intermediate reasoning quality, leading to suboptimal training results.
In this paper, we introduce \textbf{Agent Reasoning Reward Model (Agent-RRM)}, a multi-faceted reward model that produces structured feedback for agentic trajectories, including (1) an explicit reasoning trace , (2) a focused critique that provides refinement guidance by highlighting reasoning flaws, and (3) an overall score that evaluates process performance.
Leveraging these signals, we systematically investigate three integration strategies: \textbf{Reagent-C} (text-augmented refinement), \textbf{Reagent-R} (reward-augmented guidance), and \textbf{Reagent-U} (unified feedback integration).
Extensive evaluations across 12 diverse benchmarks demonstrate that Reagent-U yields substantial performance leaps, achieving 43.7\% on GAIA and 46.2\% on WebWalkerQA, validating the effectiveness of our reasoning reward model and training schemes.
Code, models, and datasets are all released to facilitate future research.

\end{abstract}
\section{Introduction}


Reinforcement Learning with Verifiable Reward (RLVR) has achieved remarkable success in improving the reasoning capabilities of Large Language Models (LLMs) \cite{liu2025understanding, feng2025onethinker, tang2025rethinking, chen2025advancing, chen2025ares}. Motivated by this progress, recent works have extended this paradigm to agents, demonstrating its potential to handle complex interactions with dynamic environments and external knowledge sources~\cite{jin2025search, wu2025webdancer, li2025websailor}.

However, previous agentic RL methods typically rely on sparse, outcome-based rewards based solely on final correctness ~\cite{jin2025search, wu2025webdancer, li2025websailor}. This design is inherently limiting for long-horizon agentic tasks requiring multi-step tool utilization~\cite{feng2025group,liu2025agentic,zhang2025rlvmr}.
In such settings, outcome-based supervision fails to differentiate high-quality intermediate reasoning from entirely incorrect attempts, for instance, treating a trajectory that fails only at the final step as a total failure. 
Consequently this coarse-grained binary supervision obscures the value of successful intermediate steps, resulting in sub-optimal performance~\cite{dong2025agentic}.


To provide more granular feedback, recent research has pivoted toward integrating Reward Models into Agentic RL. However, the effective deployment of Reward Models remains hampered by two bottlenecks. First, while step-level rewards offer finer granularity feedback~\cite{xi2025agentprm, liu2025agentic, xu2025hybrid}, they are often plagued by prohibitive annotation costs~\cite{rahman2025spark} and a susceptibility to reward hacking~\cite{zhang2025linking}. Second, existing reasoning-based Reward Models focus on pair-wise preferences~\cite{li2025one, liu2025agentic, hu2025openreward}, which frequently introduces inherent biases and fails to capture fine-grained quality gradations between trajectories or provide actionable guidance for refinement~\cite{jian2025patarm, zhang2025bradley}. 
Furthermore,  most of these efforts exclusively rely on numeric reward feedback for training, leaving the natural language critique \cite{zhang2025critique} largely unexplored, which could provide more granular guidance for agentic policy.


To this end, we develop \textbf{Agent Reasoning Reward Model (\rewardmodel{})}, a  multi-faceted evaluator designed to provide reasoning-aware feedback for agentic trajectories. 
Unlike conventional Reward Models that yield merely scalar scores or binary preferences, \rewardmodel{} conducts explicit reasoning to justify its assessments. 
For each trajectory, it generates a decomposed judgment comprising: (1) an internal reasoning trace that analyzes logical consistency of trajectory; (2) a targeted critique identifying specific flaws to guide refinement; and (3) a holistic quality score. This hierarchy of signals provides dense, multi-dimensional supervision, combining scalar rewards for global optimization with textual critiques for explicit error correction—all without necessitating ground truth.

Building upon these informative signals, we perform a systematic investigation into the integration of \rewardmodel{} and Agentic RL. We formalize this integration through a unified scheme with three variants: Text-augmented Refinement, where agents polish trajectories based on \rewardmodel{}'s textual feedback; Reward-augmented Guidance, which complements rule-based rewards with model-based signals; and Unified Feedback Integration, which harmonizes multi-source rewards with critique-augmented sampling. We denote the agent policy models of these variants as \textbf{Reagent-C}, \textbf{Reagent-R}, and \textbf{Reagent-U}, respectively. 
Notably, our experiments demonstrate that Reagent-U achieves superior performance by synthesizing these feedback modalities, reaching 43.7\% on GAIA and 46.2\% on WebWalkerQA.
Our study provides a comprehensive roadmap for harnessing multi-level feedback to accelerate agentic RL.

To support this investigation, we curate \textbf{four specialized datasets} that provide high-quality trajectories for both agent reasoning and reward model training. Extensive experiments across \textbf{12 diverse  benchmarks} demonstrate that Reagent models achieve significant performance gains, underscoring the efficacy of multi-level reasoning-based feedback signals in complex agentic tasks.

In summary, our contributions are as follows:

\begin{itemize}[topsep=2pt, itemsep=2pt, parsep=0pt]
    \item We introduce \textbf{\rewardmodel{}}, a multi-faceted evaluator that generates structured feedback including explicit reasoning rationales, actionable critiques, and holistic quality scores, providing a transparent and granular assessment.
    
    \item We systematically explore three agent variants with Agent-RRM: Text-augmented Refinement (\textbf{Reagent-C}), Reward-augmented Guidance (\textbf{Reagent-R}),  and Integrated Feedback Optimization (\textbf{Reagent-U}). This provides a roadmap for using reasoning rewards to enhance agent performance.

    \item We curate and release \textbf{four high-quality datasets} specifically tailored for training reasoning agent and reward model. These resources provide the community with valuable assets to advance research in multi-granular feedback for agentic reinforcement learning.
\end{itemize}


\section{Related Work}

\subsection{Agentic Reinforcement Learning}

Agentic Reinforcement Learning (Agentic RL) has emerged as a cornerstone for developing agents capable of operating in dynamic, open-ended environments~\cite{dong2025agentic,lu2025agentrewardbench}. 
Recent advancements~\cite{jin2025search,wang2025adatooler,xia2025agent0,wu2025reinforcing,li2025flow,song2025r1} illustrate that RL can effectively instill multi-step information-seeking and tool-use proficiencies.
For example, Search-R1~\cite{jin2025search} demonstrates that agentic RL enables LLMs to interleave multi-turn web search, substantially improving retrieval-augmented reasoning performance.
WebSailor~\cite{li2025websailor} further shows that agentic RL can scale to long-horizon web navigation, equipping agents with the ability to reduce extreme uncertainty in complex information-seeking tasks.
Agent0~\cite{xia2025agent0} introduces a co-evolutionary process where tool-aware reasoning behaviors emerge without human-curated supervision.
Despite these successes, most existing methods rely heavily on sparse, outcome-based rewards, which often limits training efficacy and hampers agent’s ability to rectify intricate intermediate errors~\cite{dong2025agentic,lin2025comprehensive}.


\subsection{Reward Modeling}

\begin{figure*}[h]
    \centering
    \includegraphics[
        width=\textwidth,
        height=0.9\textheight,
        keepaspectratio
    ]{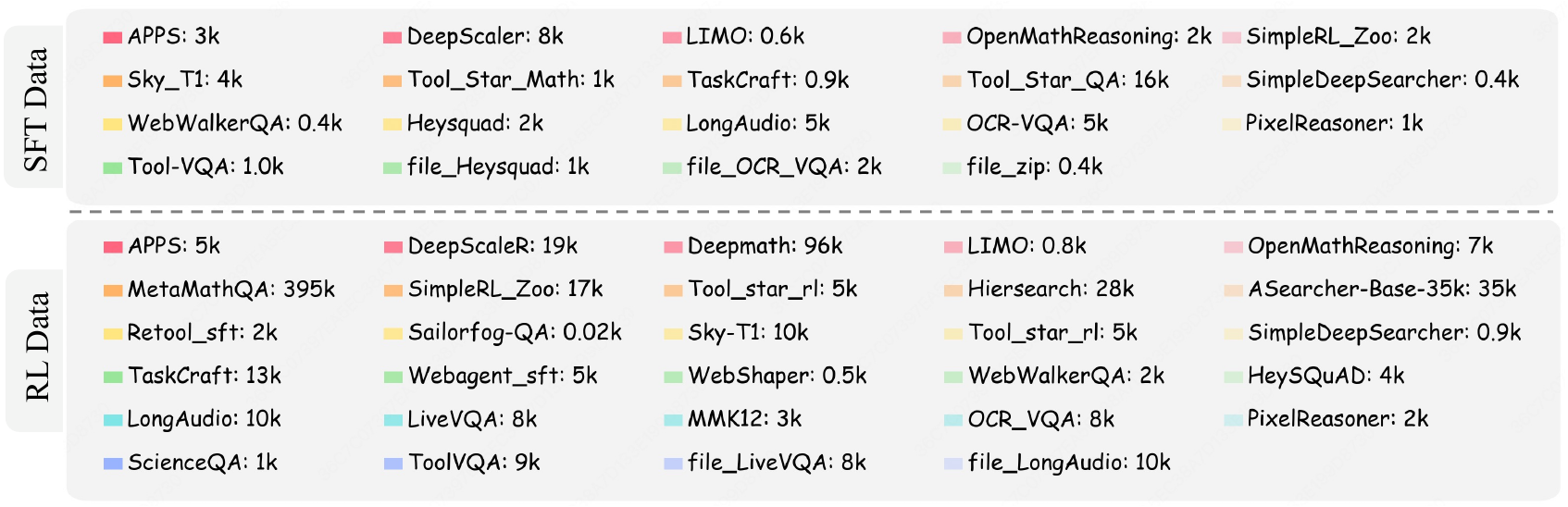}
    \caption{Detailed distribution information of  Reagent-SFT-55.6K and Reagent-RL-709K.}
    \label{fig:dataset}
\end{figure*}

Reward Models (RMs) play a central role in optimizing complex reasoning tasks by providing learning signals for policy improvement \cite{wang2024secrets,fan2025sophiavl,li2025editthinker}. Motivated by Deepseek-R1 \cite{guo2025deepseek}, reasoning-aware reward models are introduced to 
perform explicit reasoning before reward assignment for delivering higher-quality and more transparent supervision \cite{whitehouse2025j1,zhang2025r1}.
For instance, RM-R1 \cite{chen2025rm} introduces a generative reasoning-based reward model to first derive explicit reasoning rubrics and then evaluate candidate responses accordingly. 
R1-Reward \cite{zhang2025r1} proposes a multimodal reasoning reward model and introduces a stabilized RL algorithm that improves training robustness for multimodal RMs. In  agent domain, reasoning-based reward models still remain underexplored.
Atom-Searcher~\cite{deng2025atom} directly utilizes a Qwen3-30B-A3B without training as reward model to assign scores to agent steps, while PPR~\cite{xu2025hybrid} employs a process reward model to evaluate trajectory steps based on a predefined principle set. 
However, these methods remain confined to step-level scalar rewards, which are susceptible to reward hacking and fail to provide language-based guidance necessary for rectifying complex logic flaws.



\section{Method}

\subsection{Preliminaries: GRPO Framework}
\label{section:grpo}
In Group Relative Policy Optimization (GRPO)~\cite{shao2024deepseekmath}, for a query $q$ sampled from the dataset $P(q)$, the policy $\pi_\theta$ generates a group of $G$ outputs $\{o_i\}_{i=1}^G$ such that:
\vspace{-8pt}
\begin{equation}
    o_i \sim \pi_{\theta_{old}}(o|q).
\end{equation}
Let $r_i(\theta) = \frac{\pi_\theta(o_i|q)}{\pi_{\theta_{old}}(o_i|q)}$ denote the importance sampling ratio. The GRPO objective is formulated as:
\vspace{-5pt}
\begin{equation}
\begin{aligned}
    \mathcal{J}_{GRPO}&(\theta) = \mathbb{E}_{\substack{q \sim P(q)  \{o_i\} \sim \pi_{\theta_{old}}}} \\ & \bigg[ \frac{1}{G} \sum_{i=1}^G \Big(\min(r_i(\theta)A_i, \text{clip}_\epsilon)
     - \beta \mathbb{D}_{KL}^{(i)} \Big) \bigg],
\end{aligned}
\end{equation}
where $\text{clip}_\epsilon$ denotes $\text{clip}(r_i(\theta), 1-\epsilon, 1+\epsilon)A_i$, and $\mathbb{D}_{KL}^{(i)}$ denotes the KL divergence between current policy and reference model $\pi_{ref}$ for the $i$-th output:
\vspace{-5pt}
\begin{equation}
    \mathbb{D}_{KL}^{(i)} = \mathbb{D}_{KL}(\pi_\theta(o_i|q) || \pi_{ref}(o_i|q)).
\end{equation}
The advantage $A_i$ is computed by normalizing the rewards within the group $\mathbf{R} = \{R_1, \dots, R_G\}$:
\vspace{-5pt}
\begin{equation}
    A_i = \frac{R_i - \text{mean}(\mathbf{R})}{\text{std}(\mathbf{R})}.
\end{equation}
where $R_i$ is the reward for the $i$-th output assigned by the reward system.

\subsection{Agentic Tool Design}

To enable effective interaction with diverse environments, we equip the agent with a suite of six specialized tools covering information retrieval, code execution, and multi-modal perception:

\begin{figure*}[h]
    \centering
    \includegraphics[width=\linewidth]{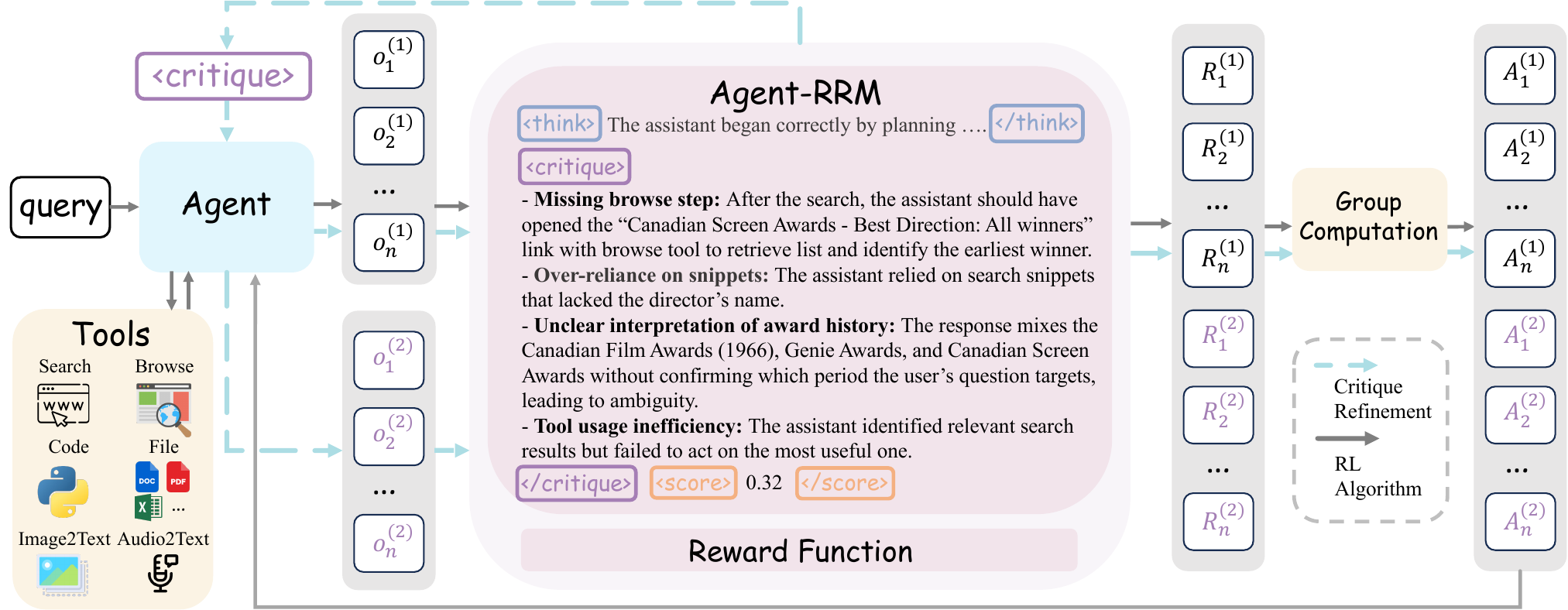}
\caption{Overview of the \textbf{Reagent} training scheme. We explore three integration variants: \textbf{Reagent-C} (\textbf{\textcolor{figblue}{blue}} arrows), \textbf{Reagent-R} (\textbf{\textcolor{figgray}{gray}} arrows), and \textbf{Reagent-U} (both arrows).}
    \label{fig:framework}
\end{figure*}

\begin{itemize}[topsep=2pt, itemsep=2pt, parsep=0pt]
    \item \textbf{Search}: Given a query, retrieve relevant search results using the Bing search engine.
    
    \item \textbf{Web Browse}: Given a URL and a query, fetch the webpage content and generate a response to the query based on the page information.
    
    \item \textbf{Python Code Interpreter}: Execute a provided Python code snippet and return the execution results.
    
    \item \textbf{File Reader}: Access and extract a file and return its textual content.
    
    \item \textbf{Image Descriptor}: Given an image and a query, generate a textual response to the query conditioned on the visual features.
    
    \item \textbf{Audio Converter}: Transcribe an input audio file into text.
\end{itemize}

\subsection{Dataset Construction}

We curate four specialized datasets spanning mathematical deduction, multimodal understanding, web-based information seeking, and complex tool utilization. These datasets support the distinct training requirements of both Reagent and \rewardmodel{}.

\paragraph{Agent Training Datasets}
We synthesize diverse QA benchmarks to enhance the agent's reasoning and tool-use capabilities. To ensure data quality, we apply a rigorous three-stage pipeline: (1) filtering samples with ambiguous ground truths; (2) cross source deduplication; and (3) difficulty-aware sampling. This yields \textbf{Reagent-RL-709K}, a comprehensive corpus of 709k question-answer pairs for RL training.
For Supervised Fine-tuning (SFT), we prioritize the holistic quality of reasoning trajectories. Using DeepSeek-V3.1, we generate and retain only trajectories that lead to correct final answers, resulting in \textbf{Reagent-SFT-55.6K} as high-quality cold-start data. Dataset distribution is shown in Figure~\ref{fig:dataset}. Specific selection criteria and filtering thresholds are detailed in Appendix~\ref{appendix:data selection}.

\paragraph{Reward Model Datasets}
Training a robust \rewardmodel{} necessitates exposure to a wide spectrum of logical error patterns and diverse response styles. Building upon the  Reagent-RL-709K, we construct two meticulously labeled datasets: \textbf{Reagent-RRM-SFT-28K} and \textbf{Reagent-RRM-RL-90K} for SFT and RL stage respectively. 
We sample reasoning trajectories from an ensemble of models including Qwen3-8B/14B, Qwen3-ARPO-DeepSearch (8B/14B), Qwen2.5-7B-ARPO, Qwen2.5-WebDancer (7B/32B), and DeepSeekV3.1 to maximize the coverage of potential error patterns. These trajectories are then annotated by GPT-OSS-120B to generate structured three-part judgments: an analytical \texttt{<think>} trace, a targeted \texttt{<critique>} of flaws, and a holistic \texttt{<score>} ($s \in [0, 1]$). 
See Appendix~\ref{appendix:reward model data} for prompts and process details.

\subsection{\rewardmodel{}: Reward Model Training}
\label{section:reward modeling}

To facilitate granular feedback, we train \textbf{Agent Reasoning Reward Model (\rewardmodel{})} to generate multi-dimensional judgments consisting of three components: \texttt{<think>}, an internal reasoning trace analyzing trajectory quality, \texttt{<critique>}, a targeted identification of reasoning or execution flaws, and \texttt{<score>}, a scalar quality assessment within $[0, 1]$. 
Following~\cite{chen2025rm,zhang2025r1}, we adopt a two-stage training procedure. First, we conduct SFT on Reagent-RRM-SFT-28K to instill the  structured output format and foundational evaluative capabilities. Subsequently, we apply GRPO on Reagent-RRM-RL-90K to refine the model's evaluative rationales and ensure the calibration of its scalar rewards. 
This training paradigm ensures that \rewardmodel{} can generate high-fidelity, self-consistent feedback even in the absence of ground-truth answers, making it highly effective for complex, open-ended agentic tasks.

\subsection{Reagent: Integrating Reasoning Rewards into Agents}

In this section, we introduce our agent policy model  \textbf{Reagent}, and present three variants that explore different ways of incorporating reasoning rewards and critiques into agentic policies.


To provide a robust starting point for RL, we fine-tune the base model on the \textbf{Reagent-SFT-55.6K} dataset. This stage ensures the agent acquires fundamental reasoning and tool-calling proficiencies. The resulting optimized policy, $\pi_{\theta_{SFT}}$, serves as the seed model for subsequent RL investigations in Section~\ref{section:paradigm i} and \ref{section:paradigm iii}.
We investigate three Reagent variants to explore the synergy between \rewardmodel{} and the agent: (1) \textbf{Textual-augmented Refinement} (\textbf{Reagent-C}), which evaluates the immediate utility of textual critiques via zero-shot, in-context refinement; (2) \textbf{Reward-augmented Guidance} (\textbf{Reagent-R}), which optimizes $\pi_{\theta_{SFT}}$ by complementing rule-based rewards with model-based scalar signals; and (3) \textbf{Unified Feedback Integration} (\textbf{Reagent-U}), which harmonizes both modalities within a joint optimization loop. The overall framework is illustrated in Figure~\ref{fig:framework}.

\subsubsection{Textual-augmented Refinement}
\label{section:paradigm ii}

This variant (\textbf{Reagent-C}) exploits textual critiques from \rewardmodel{} for training-free refinement, applied directly to the Qwen3-8B via in-context prompting.

For each query $q$, the agent first generates an initial response $o_i^{(1)} \sim \pi_{\theta}(o|q)$. Subsequently, \rewardmodel{} analyzes $o_i^{(1)}$ to produce a targeted critique $c_i$ via its \texttt{<critique>} component, identifying specific reasoning flaws or execution errors. The agent then performs a refined pass conditioned on feedback:
\vspace{-5pt}
\begin{equation}
    o_i^{(2)} \sim \pi_{\theta}(o | q, o_i^{(1)}, c_i),
    \label{eq:paradigm_ii_sampling}
\end{equation}
where the augmented context $(q, o_i^{(1)}, c_i)$ provides the original task and explicit guidance for correction. Crucially, the policy $\pi_\theta$ remains frozen in this variant, allowing us to isolate and evaluate the agent's in-context refinement capability. 
All reported results for Reagent-C correspond to the refined outputs $\{o_i^{(2)}\}$.

\subsubsection{Reward-augmented Guidance}
\label{section:paradigm i}

This variant (\textbf{Reagent-R}) utilizes the scalar score from \rewardmodel{} to provide fine-grained quality assessments of agent trajectories. Following standard GRPO sampling procedure, the agent generates $G$ outputs $o_i \sim \pi_{\theta_{old}}(o|q)$. The reward $R_i$ is defined as a combination of rule-based correctness and model-based quality evaluation:
\vspace{-5pt}
\begin{equation}
R_i = R_{rule}(q, o_i) + \lambda \cdot R_{model}(q, o_i),
\label{equation:ii reward}
\end{equation}
where $R_{rule}$ validates final answer correctness, $R_{model}$ is extracted from \rewardmodel{}'s \texttt{<score>}, and $\lambda$ is a scaling factor balancing their contributions. 
This variant alleviates the sparsity of rule-based rewards by providing reasoning-aware feedback. It enables the agent to capture a fine-grained spectrum of trajectory quality, effectively rewarding logical merit while penalizing reasoning deficiencies regardless of final answer's correctness.

\subsubsection{Unified Feedback Integration}
\label{section:paradigm iii}

This variant (\textbf{Reagent-U}) harmonizes scalar rewards and textual critique-driven refinement within a unified RL loop. By simultaneously optimizing initial generation quality and refinement capability, we investigate whether these objectives can yield synergistic improvements through mutual reinforcement for agent. 

For each query $q$, the agent performs a two-stage sampling:
\vspace{-5pt}
\begin{equation}
    o_i^{(1)} \sim \pi_{\theta_{old}}(o|q), \quad o_i^{(2)} \sim \pi_{\theta_{old}}(o|q, o_i^{(1)}, c_i),
\end{equation}
where $o_i^{(1)}$ is the initial attempt and $o_i^{(2)}$ is the refined response guided by \texttt{<critique>} $c_i$ generated by \rewardmodel{}. 

We pool all trajectories from both stages into $\mathcal{G}_{pool} = \{o_i^{(k)} \mid i \in [G], k \in \{1, 2\}\}$ and compute combined reward $R_i^{(k)}$ via Eq.~\ref{equation:ii reward}. The advantage is computed across this unified pool:
\vspace{-5pt}
\begin{equation}
    A_i^{(k)} = \frac{R_i^{(k)} - \text{mean}(\mathbf{R}_{pool})}{\text{std}(\mathbf{R}_{pool})},
\end{equation}
where $\mathbf{R}_{pool} = \{R_i^{(k)} \mid o_i^{(k)} \in \mathcal{G}_{pool}\}$. The unified objective is formulated as:
\vspace{-5pt}
\begin{equation}
\small
\begin{aligned}
    \mathcal{J}_{U}(\theta) = \mathbb{E} \bigg[ \frac{1}{2G} \sum_{k=1}^2 \sum_{i=1}^G & \Big( \min(r_i^{(k)}(\theta)A_i^{(k)}, \text{clip}_\epsilon) \\
    & - \beta \mathbb{D}_{KL}^{(i,k)} \Big) \bigg],
\end{aligned}
\end{equation}
where the importance ratio $r_i^{(k)}(\theta)$ and KL penalty $\mathbb{D}_{KL}^{(i,k)}$ are computed relative to their respective contexts. By normalizing advantages across all initial and refined trajectories, Reagent-U encourages the agent to optimize for overall trajectory quality, effectively boosting the agent's core reasoning and tool-calling performance.
Notably, textual critiques are utilized exclusively during the training phase to internalize reasoning capabilities; at inference time, ReAgent-U operates as a standard agent without additional critique refinement or external guidance.

\begin{table*}[ht]
\centering
\caption{Comprehensive Evaluation on General Agent and Search Benchmarks.}
\label{tab:comprehensive_performance}
\resizebox{1\linewidth}{!}{%
    \setlength{\tabcolsep}{1.1mm}
    \renewcommand\arraystretch{1.3}
\begin{tabular}{ll cccc c cccc c c c c}
\hline
& & \multicolumn{4}{c}{\textbf{GAIA (text)}} && \multicolumn{4}{c}{\textbf{WebWalkerQA}} && \textbf{HLE} && \textbf{xbench} \\
\cmidrule{3-6} \cmidrule{8-11} \cmidrule{13-13} \cmidrule{15-15}
\textbf{Method} & \textbf{Backbone} & Lv.1 & Lv.2 & Lv.3 & \textbf{Avg.} && Easy & Med. & Hard & \textbf{Avg.} && \textbf{Avg.} && \textbf{Avg.} \\
\hline
\rowcolor{gray!10} \multicolumn{15}{l}{\textit{Proprietary Agents}} \\
- & OpenAI-o3 & - & - & - & 70.5 && - & - & - & 71.7 && 20.2 && 66.0 \\
- & o1-preview & - &-& -& - && 11.9 & 10.4 &7.9&9.9 && 11.1 && - \\
- & Calude-4-Sonnet & - & - & - & 68.3 && - & - & - & 61.7 && 20.2 && 64.0 \\
- & OpenAI DeepResearch & - & - & - & 67.4 && - & - & - & - && 26.6 && - \\
\hline
\rowcolor{gray!10} \multicolumn{15}{l}{\textit{Open-source Baselines ($\leq$8B)}} \\
WebThinker~\cite{li2025webthinker} & Qwen3-8B & 43.6 & 11.5 & 0.0 & 22.3 && 6.7 & 13.1 & 16.9 & 13.0 && 6.6 && 13.0 \\
WebDancer~\cite{wu2025webdancer} & Qwen2.5-7B & 41.0 & 30.7 & 0.0 & 31.0 && 40.6 & 44.1 & 28.2 & 36.0 && - && - \\
VerlTool~\cite{jiang2025verltool} & Qwen3-8B & - & - & - & 34.0 && - & - & - & - && 8.4 && - \\
ARPO~\cite{dong2025agentic} & Qwen3-8B & 53.9 & 32.7 & \textbf{16.7} & 38.8 && 26.7 & 33.3 & 29.6 & 30.5 && 8.8 && 25.0 \\
\hline
\rowcolor{gray!10} \multicolumn{15}{l}{\textit{Open-source Baselines ($\leq$32B)}} \\
- & QwQ-32B & 30.9 & 6.5 & 5.2 & 18.9 && 7.5 & 2.1 & 4.2 & 3.8 && 6.4 && 10.0 \\
- & DeepSeek-R1-671B & 40.5 & 21.2 & 5.2 & 25.2 && 5.0 & 11.8 & 11.3 & 10.0 && 8.6 && 32.0 \\
Tree-GRPO~\cite{ji2025tree} & Qwen2.5-14B & 20.8 & 24.3 & 7.3 & 21.0 && 11.1 & 15.5 & 10.8 &12.8 &&- && -\\
ARPO~\cite{dong2025agentic} & Qwen3-14B & \underline{56.4} & \underline{40.4} & \textbf{16.7} & \textbf{43.7} && 31.1 & 42.9 & 31.0 & 36.0 && \underline{10.0} && 32.0 \\
Search-o1~\cite{li2025search} & QwQ-32B-Preview & 53.8 & 34.6 & \textbf{16.7} & 39.8 && 43.1 & 35.0 & 27.1 & 34.1 && \textbf{10.8} && 40.0 \\
WebDancer~\cite{wu2025webdancer} & Qwen2.5-32B & 46.1 & \textbf{44.2} & \underline{8.3} & \underline{40.7} && 44.3 & \underline{46.7} & 29.2 & 38.4 && - && 38.0 \\
\hline
\rowcolor{gray!10} \multicolumn{15}{l}{\textit{Open-source Baselines with Process Reward}} \\
Atom-Searcher~\cite{deng2025atom} & Qwen2.5-7B & 18.0& 21.2 & 0.0 & 17.5 && 31.7 & 23.7 & 37.0 & 27.9 && \underline{10.0} && 21.0 \\
\hline
\rowcolor{gray!10} \multicolumn{15}{l}{\textit{Our Agents}} \\
- & Qwen3-8B & 28.2 & 21.2 & 0.0 & 21.4 && 31.1 & 28.6 & 28.2 & 29.0 && 4.0 && 9.0 \\
Reagent w/o Agent-RRM & Qwen3-8B & 41.0 & 36.5 & 0.0 & 34.0 && 44.4& 45.0 & 41.3 & 43.5 && 6.8 && 32.0 \\
\rowcolor{mypurple} Reagent-C (Direct Inference)& Qwen3-8B  & 30.8 & 23.1 & \textbf{16.7} & 25.2 && 35.6&38.1 & 32.4 & 35.5 &&  4.6 && 15.0 \\
\rowcolor{mypurple} Reagent-R & Qwen3-8B & 51.3 &30.8 & \textbf{16.7} & 36.9 && \underline{47.5} & 46.0 & \underline{42.9} & \underline{45.3} && \underline{10.0} && \underline{41.0} \\
\rowcolor{mypurple} Reagent-U & Qwen3-8B & \textbf{59.0} & 38.5 & \textbf{16.7} & \textbf{43.7} && \textbf{49.2} & \textbf{46.8} & \textbf{43.3} & \textbf{46.2} && \textbf{10.8} && \textbf{43.0} \\
\hline
\end{tabular}}
\end{table*}

\section{Experiments}

\paragraph{Benchmarks} We evaluate comprehensively on multiple challenging benchmarks. \textbf{(1) Mathematical Reasoning: }AIME24~\cite{aime24}, AIME25~\cite{aime25}, GSM8K~\cite{cobbe2021gsm8k} and MATH500~\cite{lightman2023let}.\textbf{(2) Knowledge-Intensive Reasoning: } HotpotQA~\cite{yang2018hotpotqa}, 2Wiki~\cite{xanh2020_2wikimultihop}, Bamboogle~\cite{press2023measuring} and MuSiQue~\cite{trivedi2021musique}. \textbf{(3) General Agent and Search Reasoning: } GAIA~\cite{mialon2023gaia}, WebWalkerQA~\cite{wu2025webwalker}, Humanity's Last Exam (HLE)~\cite{phan2025humanity} and xbench~\cite{chen2025xbench}.

\paragraph{Implementation Details}


Following \cite{ dong2025agentic,feng2025video,wu2025webdancer}, we employ a two-phase training pipeline: Supervised Fine-Tuning followed by Reinforcement Learning. This training protocol mitigates optimization instability in early RL stages and equips the agent with the foundational skills necessary for effective tool interaction. 
Both agent models and reward model are initialized from Qwen3-8B~\cite{yang2025qwen3}.

Both our agent models and Agent-RRM are trained on 8 NVIDIA A800-80G GPUs.
Batch size is set to 32 for both SFT and RL. We use AdamW optimizer. Learning rate is set to $1\times 10^{-5}$ for SFT and $5\times 10^{-7}$ for RL. $\lambda$ in Eq.~\ref{equation:ii reward} is set to 0.3.
Detailed hyperparameters and compute resources are deferred to Appendix~\ref{appendix:training}.
For evaluation metrics, following \cite{dong2025agentic}, we utilize Qwen2.5-72B-Instruct as judge model to perform binary scoring based on ground truth and agent prediction. To ensure a rigorous comparison with prior works~\cite{dong2025agentic,wu2025webdancer}, unless otherwise specified, we report pass@1 using a decoding temperature of 0.6 and top-p of 0.95.  Evaluation details are shown in Appendix~\ref{appendix:eval}.

\begin{table*}[h]
\centering
\caption{Results on Knowledge-Intensive Reasoning and Math Benchmarks. (HQA: HotpotQA)}
\resizebox{1\linewidth}{!}{%
    \setlength{\tabcolsep}{1.2mm}
    \renewcommand\arraystretch{1.35}
\begin{tabular}{ll cccc c cccc}
\hline
& & \multicolumn{4}{c}{\textbf{Knowledge-Intensive Reasoning}} && \multicolumn{4}{c}{\textbf{Mathematical Reasoning}} \\
\cmidrule{3-6} \cmidrule{8-11}
\textbf{Method} & \textbf{Backbone} & \textbf{HQA} & \textbf{2Wiki} & \textbf{Bamboogle} & \textbf{MuSiQue} && \textbf{AIME24} & \textbf{AIME25} & \textbf{MATH500} & \textbf{GSM8K} \\
\hline
\rowcolor{gray!10} \multicolumn{11}{l}{\textit{Proprietary Agents}} \\
- & GPT-4o & 54.0 & 49.5 & 68.8 & 24.0 && 13.4 & 25.7 & 60.3 & - \\
- & o1-preview & -&- &- &- && 46.7 & - & 85.5 & -\\
- & Claude-4-Sonnet & -&-&-&-&&43.4 & 33.1& 93.4 &-\\
\hline
\rowcolor{gray!10} \multicolumn{11}{l}{\textit{Open-source Baselines ($\leq$8B)}} \\
Search-R1~\cite{jin2025search} & Qwen2.5-7B & 43.3 & 38.2 & 43.2 & 19.6 && - & - & - & - \\
VerlTool~\cite{jiang2025verltool} & Qwen2.5-7B\protect\footnotemark & 42.6 & 39.2 & 38.4 & 18.0 && 36.7 & 33.3 & 82.8 & 92.1 \\
ARPO~\cite{dong2025agentic} & Qwen2.5-7B & 58.8 & 76.1 & 71.5 & \underline{31.1} && 33.3 & 30.0 & 88.8 & 92.2 \\
ARPO~\cite{dong2025agentic} & Qwen3-8B & - & - & - & - && 33.3 & 30.0 & 88.4 & 93.4 \\
AgentFlow~\cite{li2025flow} & Qwen2.5-7B & 57.0 & 77.2 & 69.6 & 25.3 && 40.0 & - & - & - \\

\hline
\rowcolor{gray!10} \multicolumn{11}{l}{\textit{Open-source Baselines ($\leq$32B)}} \\
Search-o1~\cite{li2025search} & QwQ-32B-Preview & 45.2 & 58.0 & 56.0 & 16.6 && \underline{56.7} & - & 86.4 & - \\
Tree-GRPO~\cite{ji2025tree} & Qwen2.5-14B & 50.2 & 50.5 & 54.4 & 25.9 && - & - & - & - \\
ARPO~\cite{dong2025agentic} & Qwen3-14B & -& - & - & - && 36.7 & 30.0 & 83.0 & 93.6 \\
\hline
\rowcolor{gray!10} \multicolumn{11}{l}{\textit{Open-source Baselines with Process Reward}} \\
Atom-Searcher~\cite{deng2025atom} & Qwen2.5-7B & 57.3 & 66.9 & 70.7 & 27.6 && - & - & - & - \\
PPR-Instruct~\cite{xu2025hybrid} & Qwen2.5-7B & 38.7 & 31.0 & 41.2 & 15.5 && - & - & - & - \\
\hline
\rowcolor{gray!10} \multicolumn{11}{l}{\textit{Our Agents}} \\
- & Qwen3-8B & 52.0 & 58.0 & 53.6 & 22.1 && 46.7 & 40.0 & 90.4 & 94.6 \\
Reagent w/o Agent-RRM & Qwen3-8B & 65.8 & 77.0& 61.6 & 28.1 && 50.0 & 43.3 & 90.8 & 94.5 \\
\rowcolor{mypurple} Reagent-C (Direct Inference)& Qwen3-8B & 61.0 & 68.9 & 61.6 & 25.0 && \underline{56.7} & \underline{46.7} & \textbf{93.8} & \underline{94.9} \\
\rowcolor{mypurple} Reagent-R & Qwen3-8B & \underline{67.9} & \textbf{79.0} & \underline{72.8} & 28.3 && 53.3 & \textbf{50.0} & \underline{92.2} & 94.1 \\
\rowcolor{mypurple} Reagent-U & Qwen3-8B & \textbf{68.1} & \underline{78.8} & \textbf{76.8} & \textbf{31.3} && \textbf{60.0} & \textbf{50.0} & \textbf{93.8} & \textbf{95.1} \\
\hline
\end{tabular}}
\vspace{0.03in}
\label{tab:combined_results}
\end{table*}

\subsection{Can Textual Critiques Guide Inference-Time Refinement?}
\label{ablation:critique}

To investigate the direct impact of textual critiques, we evaluate Reagent-C--a training-free variant--on Qwen3-8B using Agent-RRM for critique guidance. 
As shown in Table~\ref{tab:comprehensive_performance} and Table~\ref{tab:combined_results}, Reagent-C achieves consistent performance gains across all benchmarks without any parameter updates. Improvements are particularly pronounced in Mathematical Reasoning, while solid advancements are also observed in General Agentic and Knowledge-Intensive tasks. We attribute this versatility to Agent-RRM’s diagnostic capacity, which effectively pinpoints logical fallacies and tool-execution errors within complex trajectories. See case study in Appendix~\ref{appendix:case study}.

The results confirm that the second response $\{o_i^{(2)}\}$ consistently achieves better performance compared to the initial response $\{o_i^{(1)}\}$.
Crucially, the widening margin between the first and second response underscores that many initial failures stem from transient execution errors or logical oversights. Our critiques are uniquely positioned to rectify these flaws by offering precise, actionable feedback. Furthermore, since \rewardmodel{} operates without access to ground-truth answers, these performance gains empirically validate its capacity to diagnose reasoning flaws and tool-execution errors. This highlights that textual critiques offer the high-granularity supervision essential for mastering complex, multi-step agentic tasks.


\footnotetext{\textit{VerlTool} backbones: Qwen2.5-7B (knowledge) and Qwen2.5-Math-7B (math).}

\subsection{Does Model-based Reward Improve Learning?}
\label{ablation:model reward}

To explore whether dense model-based rewards can alleviate reward sparsity in agentic RL, 
we evaluate Reagent-R, which augments rule-based outcome rewards with holistic reasoning-level scores from \rewardmodel{}. 
As shown in Table~\ref{tab:comprehensive_performance} and Table~\ref{tab:combined_results}, Reagent-R consistently outperforms rule-based reward baseline (Reagent w/o Agent-RRM) across all benchmarks.
Specifically, Reagent-R achieves 72.8\% on Bamboogle and 41.0\% on xbench, surpassing Reagent w/o Agent-RRM by 11.2 and 9.0 percentage points, respectively.
These results suggest that holistic model-based rewards provide more informative feedback for complex, multi-step reasoning scenarios, where sparse binary outcomes often provide overly coarse and limited guidance for learning.

Reagent-R serves as a critical ablation to isolate the impact of scalar supervision by excluding the textual critiques used in Reagent-U. While Reagent-R consistently outperforms sparse-reward baselines, it remains inferior to Reagent-U across most tasks. This performance gap suggests that while continuous scores better differentiate trajectory quality, they lack the explicit, structural guidance inherent in textual feedback.
This indicates a clear need for supervision with richer informational granularity, such as the textual critiques integrated into Reagent-U.

\subsection{Does Unified Feedback Synergistically Boost Performance?}
\label{experiment:unified training}

Evaluations in Tables~\ref{tab:comprehensive_performance} and \ref{tab:combined_results} reveal that the unified feedback mechanism in Reagent-U consistently outperforms all baselines across a diverse spectrum of reasoning and agentic benchmarks. 
Specifically, Reagent-U achieves 43.7\% on GAIA (text) and 46.2\% on WebWalkerQA, surpassing all compared methods. 
Beyond its excellence in general agentic and search tasks, Reagent-U maintains robustness in knowledge-intensive and mathematical reasoning, securing 76.8\% on Bamboogle and 60.0\% on AIME24. In contrast, many existing baselines fail to generalize across diverse domains, often suffering from significant performance trade-offs.
This balanced proficiency indicates that Reagent-U augments multi-tool, multi-turn reasoning capabilities rather than merely optimizing for web search. Such results demonstrate a comprehensive long-horizon decision-making ability, effectively showing that the unified feedback mechanism—integrating both scalar rewards and textual critiques—allows the agent to internalize a more sophisticated policy across complex, heterogeneous tasks.

\subsection{Beyond Text-Only: Cross-Modal Reasoning and Complex Tool Use} 

\begin{table}[t]
\centering
\caption{Performance of Reagent-U on GAIA text set and full set (multi-modal). }
\resizebox{\linewidth}{!}{%
    \setlength{\tabcolsep}{1.2mm}
    \renewcommand\arraystretch{1.6}
\begin{tabular}{lcccc}
\hline
\textbf{Model} & 
\multicolumn{2}{c}{\textbf{GAIA (text)}} & 
\multicolumn{2}{c}{\textbf{GAIA (full)}}  \\
\cline{2-3} \cline{4-5}
& \textbf{pass@1} & \textbf{pass@3} & \textbf{pass@1} & \textbf{pass@3} \\
\hline
Qwen3-8B~\cite{yang2025qwen3}  & 21.4 & 24.3 & 20.0 & 26.7 \\
MCP-R1~\cite{anonymous2025mcpr}       & 39.8 & 52.4 & 37.6 & 51.5 \\
\rowcolor{mypurple} Reagent-U & \textbf{43.7} & \textbf{53.4} & \textbf{38.8} & \textbf{53.9} \\
\hline
\end{tabular}}
\label{tab:gaia}
\end{table}

To evaluate Reagent-U's proficiency across diverse modalities and tools, we conduct analysis on full GAIA benchmark. 
While existing studies~\cite{dong2025agentic,jiang2025verltool,li2025websailor,li2025flow} focus on the \texttt{text} subset, which emphasizes web navigation and information retrieval ability,
we argue that such a narrow scope overlooks the heterogeneous reasoning capabilities required for complex real-world tasks. 
By evaluating on the GAIA \texttt{full} set, we challenge the agent with tasks requiring the integration of open-domain search, multimodal interpretation, python coding, and file-based reasoning. As shown in Table~\ref{tab:gaia}, Reagent-U not only maintains competitive performance on the text subset but also significantly outperforms baselines on the full set. These results confirm that Reagent-U fosters a versatile agentic intelligence that generalizes across a broad task spectrum rather than overfitting to specific text-based requirements.

\begin{figure}
    \centering
    \includegraphics[width=\linewidth]{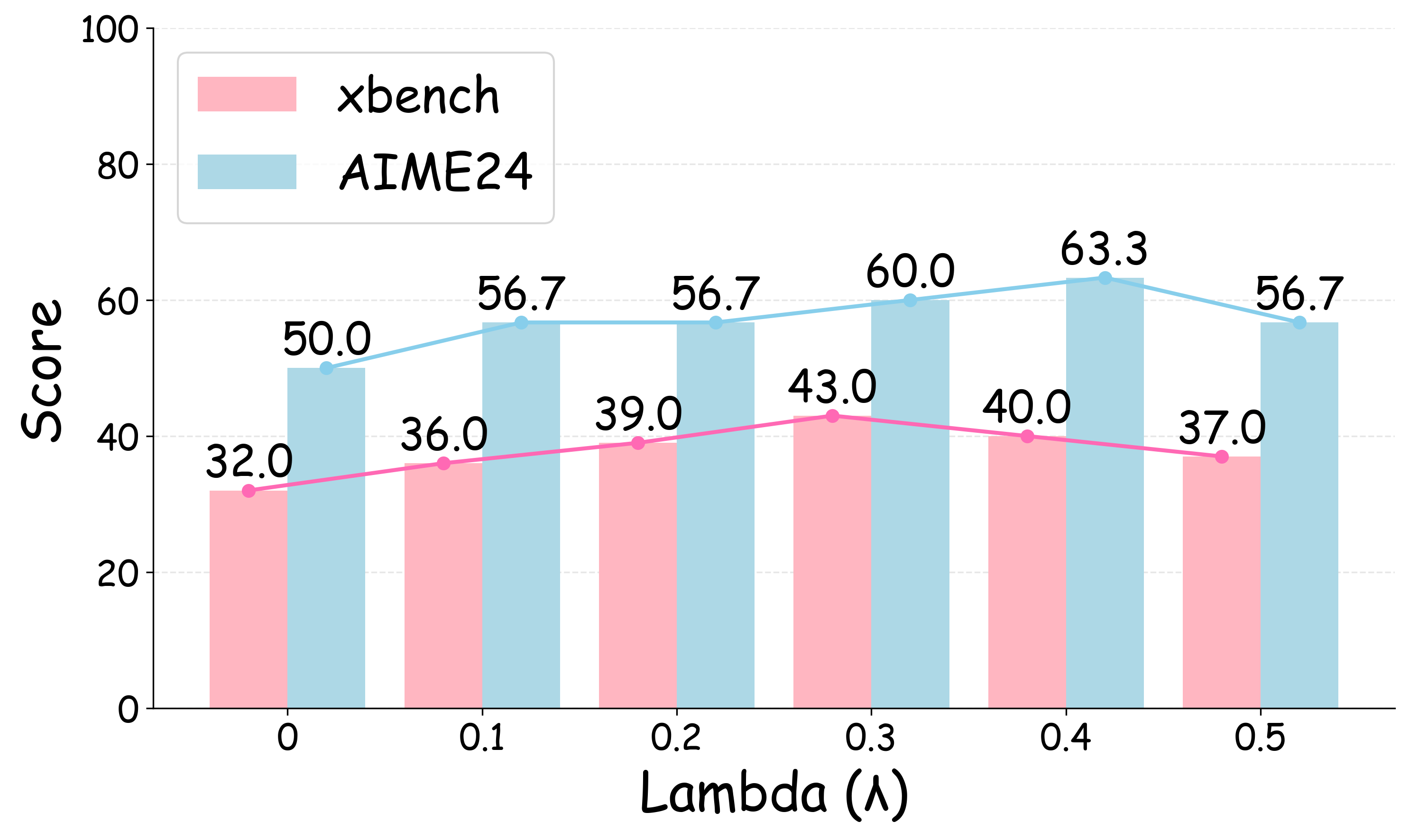}
    \vspace{-0.3in}
    \caption{Impact of Agent-RRM reward weight $\lambda$ on task performance. }
    \label{fig:parameter analysis}
    \vspace{-0.1in}
\end{figure}

\subsection{Parameter Analysis on $\lambda$}

To evaluate the impact of the Agent-RRM reward weight $\lambda$, we conduct a parameter analysis on AIME24 (math) and xbench (deep search). 
Figure~\ref{fig:parameter analysis} shows that agent performance initially increases with rising $\lambda$ values, demonstrating that the integration of reasoning-based rewards enhances the agent's decision-making compared to the baseline ($\lambda=0$). 
Specifically, performance reaches a plateau between $\lambda \in [0.2, 0.4]$, followed by a slight decline at $\lambda=0.5$. This trend suggests that while moderate reasoning feedback provides essential supervisory signals, a disproportionately high weight may over-emphasize intermediate steps at the expense of final task completion. Consequently, balancing Agent-RRM rewards with rule-based outcome reward is crucial to maintain an optimal trade-off between reasoning and outcome supervision.
\section{Conclusion}

In this work, we introduce Agent-RRM, a multi-faceted reasoning reward model designed to provide textual critiques and holistic reasoning-aware reward.
Building upon this, we present Reagent, a comprehensive scheme designed to explore the efficacy of multi-dimensional feedback in agentic learning. 
Our systematic evaluations reveal that while textual critiques provide diagnostic guidance for inference-time refinement, model-based rewards serve to mitigate signal sparsity during training. Together, these signals significantly bolster the agent's long-horizon reasoning and multi-step tool-use proficiency, leading to consistent gains across diverse complex, multi-modal tasks.

\section{Limitations}


We discuss the limitations of our work and potential directions for future research as follows:

First, our current experiments primarily focus on models at the 8B parameter scale. While this setting demonstrates the efficacy of our Reagent scheme, its scaling behavior on larger-scale models remains to be explored. Future work could investigate how more powerful base models might further amplify the benefits of structured reasoning feedback. Second, moving beyond standardized benchmarks to handle broader toolsets and more intricate reasoning chains is essential. Future works can explore open-ended, real-world applications (e.g., AI for science) that involve more diverse toolsets and unpredictable task environments to further validate the scheme's adaptability.

\bibliography{acl_latex}

@article{li2025search,
  title={Search-o1: Agentic search-enhanced large reasoning models},
  author={Li, Xiaoxi and Dong, Guanting and Jin, Jiajie and Zhang, Yuyao and Zhou, Yujia and Zhu, Yutao and Zhang, Peitian and Dou, Zhicheng},
  journal={arXiv preprint arXiv:2501.05366},
  year={2025}
}

@article{dong2025agentic,
  title={Agentic reinforced policy optimization},
  author={Dong, Guanting and Mao, Hangyu and Ma, Kai and Bao, Licheng and Chen, Yifei and Wang, Zhongyuan and Chen, Zhongxia and Du, Jiazhen and Wang, Huiyang and Zhang, Fuzheng and others},
  journal={arXiv preprint arXiv:2507.19849},
  year={2025}
}

@article{wu2025webdancer,
  title={Webdancer: Towards autonomous information seeking agency},
  author={Wu, Jialong and Li, Baixuan and Fang, Runnan and Yin, Wenbiao and Zhang, Liwen and Tao, Zhengwei and Zhang, Dingchu and Xi, Zekun and Fu, Gang and Jiang, Yong and others},
  journal={arXiv preprint arXiv:2505.22648},
  year={2025}
}

@article{li2025websailor,
  title={WebSailor: Navigating Super-human Reasoning for Web Agent},
  author={Li, Kuan and Zhang, Zhongwang and Yin, Huifeng and Zhang, Liwen and Ou, Litu and Wu, Jialong and Yin, Wenbiao and Li, Baixuan and Tao, Zhengwei and Wang, Xinyu and others},
  journal={arXiv preprint arXiv:2507.02592},
  year={2025}
}

@article{chen2025advancing,
  title={Advancing Multimodal Reasoning: From Optimized Cold Start to Staged Reinforcement Learning},
  author={Chen, Shuang and Guo, Yue and Su, Zhaochen and Li, Yafu and Wu, Yulun and Chen, Jiacheng and Chen, Jiayu and Wang, Weijie and Qu, Xiaoye and Cheng, Yu},
  journal={arXiv preprint arXiv:2506.04207},
  year={2025}
}

@article{chen2025ares,
  title={ARES: Multimodal Adaptive Reasoning via Difficulty-Aware Token-Level Entropy Shaping},
  author={Chen, Shuang and Guo, Yue and Ye, Yimeng and Huang, Shijue and Hu, Wenbo and Li, Haoxi and Zhang, Manyuan and Chen, Jiayu and Guo, Song and Peng, Nanyun},
  journal={arXiv preprint arXiv:2510.08457},
  year={2025}
}

@article{deng2025atom,
  title={Atom-searcher: Enhancing agentic deep research via fine-grained atomic thought reward},
  author={Deng, Yong and Wang, Guoqing and Ying, Zhenzhe and Wu, Xiaofeng and Lin, Jinzhen and Xiong, Wenwen and Dai, Yuqin and Yang, Shuo and Zhang, Zhanwei and Wang, Qiwen and others},
  journal={arXiv preprint arXiv:2508.12800},
  year={2025}
}

@article{jin2025search,
  title={Search-r1: Training llms to reason and leverage search engines with reinforcement learning},
  author={Jin, Bowen and Zeng, Hansi and Yue, Zhenrui and Yoon, Jinsung and Arik, Sercan and Wang, Dong and Zamani, Hamed and Han, Jiawei},
  journal={arXiv preprint arXiv:2503.09516},
  year={2025}
}

@article{jiang2025verltool,
  title={Verltool: Towards holistic agentic reinforcement learning with tool use},
  author={Jiang, Dongfu and Lu, Yi and Li, Zhuofeng and Lyu, Zhiheng and Nie, Ping and Wang, Haozhe and Su, Alex and Chen, Hui and Zou, Kai and Du, Chao and others},
  journal={arXiv preprint arXiv:2509.01055},
  year={2025}
}

@article{ji2025tree,
  title={Tree search for llm agent reinforcement learning},
  author={Ji, Yuxiang and Ma, Ziyu and Wang, Yong and Chen, Guanhua and Chu, Xiangxiang and Wu, Liaoni},
  journal={arXiv preprint arXiv:2509.21240},
  year={2025}
}

@misc{aime24,
      title={American Invitational Mathematics Examination (AIME) 2024}, 
      author={Zhang, Yifan and Math-AI, Team},
      year={2024},
}

@misc{aime25,
      title={American Invitational Mathematics Examination (AIME) 2025}, 
      author={Zhang, Yifan and Math-AI, Team},
      year={2025},
}

@article{cobbe2021gsm8k,
  title={Training Verifiers to Solve Math Word Problems},
  author={Cobbe, Karl and Kosaraju, Vineet and Bavarian, Mohammad and Chen, Mark and Jun, Heewoo and Kaiser, Lukasz and Plappert, Matthias and Tworek, Jerry and Hilton, Jacob and Nakano, Reiichiro and Hesse, Christopher and Schulman, John},
  journal={arXiv preprint arXiv:2110.14168},
  year={2021}
}

@inproceedings{lightman2023let,
  title={Let's verify step by step},
  author={Lightman, Hunter and Kosaraju, Vineet and Burda, Yuri and Edwards, Harrison and Baker, Bowen and Lee, Teddy and Leike, Jan and Schulman, John and Sutskever, Ilya and Cobbe, Karl},
  booktitle={The Twelfth International Conference on Learning Representations},
  year={2023}
}

@inproceedings{yang2018hotpotqa,
  title={{HotpotQA}: A Dataset for Diverse, Explainable Multi-hop Question Answering},
  author={Yang, Zhilin and Qi, Peng and Zhang, Saizheng and Bengio, Yoshua and Cohen, William W. and Salakhutdinov, Ruslan and Manning, Christopher D.},
  booktitle={Conference on Empirical Methods in Natural Language Processing ({EMNLP})},
  year={2018}
}

@inproceedings{xanh2020_2wikimultihop,
    title = "Constructing A Multi-hop {QA} Dataset for Comprehensive Evaluation of Reasoning Steps",
    author = "Ho, Xanh  and
      Duong Nguyen, Anh-Khoa  and
      Sugawara, Saku  and
      Aizawa, Akiko",
    booktitle = "Proceedings of the 28th International Conference on Computational Linguistics",
    month = dec,
    year = "2020",
    address = "Barcelona, Spain (Online)",
    publisher = "International Committee on Computational Linguistics",
    url = "https://www.aclweb.org/anthology/2020.coling-main.580",
    pages = "6609--6625",
}

@inproceedings{press2023measuring,
  title={Measuring and narrowing the compositionality gap in language models},
  author={Press, Ofir and Zhang, Muru and Min, Sewon and Schmidt, Ludwig and Smith, Noah A and Lewis, Mike},
  booktitle={Findings of the Association for Computational Linguistics: EMNLP 2023},
  pages={5687--5711},
  year={2023}
}

@article{trivedi2021musique,
  title={{M}u{S}i{Q}ue: Multihop Questions via Single-hop Question Composition},
  author={Trivedi, Harsh and Balasubramanian, Niranjan and Khot, Tushar and Sabharwal, Ashish},
  journal={Transactions of the Association for Computational Linguistics},
  year={2022},
  publisher={MIT Press}
}

@article{wu2025webwalker,
  title={Webwalker: Benchmarking llms in web traversal},
  author={Wu, Jialong and Yin, Wenbiao and Jiang, Yong and Wang, Zhenglin and Xi, Zekun and Fang, Runnan and Zhang, Linhai and He, Yulan and Zhou, Deyu and Xie, Pengjun and others},
  journal={arXiv preprint arXiv:2501.07572},
  year={2025}
}

@inproceedings{mialon2023gaia,
  title={Gaia: a benchmark for general ai assistants},
  author={Mialon, Gr{\'e}goire and Fourrier, Cl{\'e}mentine and Wolf, Thomas and LeCun, Yann and Scialom, Thomas},
  booktitle={The Twelfth International Conference on Learning Representations},
  year={2023}
}

@article{phan2025humanity,
  title={Humanity's last exam},
  author={Phan, Long and Gatti, Alice and Han, Ziwen and Li, Nathaniel and Hu, Josephina and Zhang, Hugh and Zhang, Chen Bo Calvin and Shaaban, Mohamed and Ling, John and Shi, Sean and others},
  journal={arXiv preprint arXiv:2501.14249},
  year={2025}
}

@misc{rllm2025,
  title={rLLM: A Framework for Post-Training Language Agents},
  author={Sijun Tan and Michael Luo and Colin Cai and Tarun Venkat and Kyle Montgomery and Aaron Hao and Tianhao Wu and Arnav Balyan and Manan Roongta and Chenguang Wang and Li Erran Li and Raluca Ada Popa and Ion Stoica},
  year={2025},
  howpublished={\url{https://pretty-radio-b75.notion.site/rLLM-A-Framework-for-Post-Training-Language-Agents-21b81902c146819db63cd98a54ba5f31}},
  note={Notion Blog},
  year={2025}
}

@article{song2025r1,
  title={R1-searcher: Incentivizing the search capability in llms via reinforcement learning},
  author={Song, Huatong and Jiang, Jinhao and Min, Yingqian and Chen, Jie and Chen, Zhipeng and Zhao, Wayne Xin and Fang, Lei and Wen, Ji-Rong},
  journal={arXiv preprint arXiv:2503.05592},
  year={2025}
}

@article{feng2025group,
  title={Group-in-group policy optimization for llm agent training},
  author={Feng, Lang and Xue, Zhenghai and Liu, Tingcong and An, Bo},
  journal={arXiv preprint arXiv:2505.10978},
  year={2025}
}

@article{liu2025agentic,
  title={Agentic Reinforcement Learning with Implicit Step Rewards},
  author={Liu, Xiaoqian and Wang, Ke and Wu, Yuchuan and Huang, Fei and Li, Yongbin and Zhang, Junge and Jiao, Jianbin},
  journal={arXiv preprint arXiv:2509.19199},
  year={2025}
}

@inproceedings{zheng2024llamafactory,
  title={LlamaFactory: Unified Efficient Fine-Tuning of 100+ Language Models},
  author={Yaowei Zheng and Richong Zhang and Junhao Zhang and Yanhan Ye and Zheyan Luo and Zhangchi Feng and Yongqiang Ma},
  booktitle={Proceedings of the 62nd Annual Meeting of the Association for Computational Linguistics (Volume 3: System Demonstrations)},
  address={Bangkok, Thailand},
  publisher={Association for Computational Linguistics},
  year={2024},
  url={http://arxiv.org/abs/2403.13372}
}

@article{sheng2024hybridflow,
  title   = {HybridFlow: A Flexible and Efficient RLHF Framework},
  author  = {Guangming Sheng and Chi Zhang and Zilingfeng Ye and Xibin Wu and Wang Zhang and Ru Zhang and Yanghua Peng and Haibin Lin and Chuan Wu},
  year    = {2024},
  journal = {arXiv preprint arXiv: 2409.19256}
}

@article{li2025flow,
  title={In-the-flow agentic system optimization for effective planning and tool use},
  author={Li, Zhuofeng and Zhang, Haoxiang and Han, Seungju and Liu, Sheng and Xie, Jianwen and Zhang, Yu and Choi, Yejin and Zou, James and Lu, Pan},
  journal={arXiv preprint arXiv:2510.05592},
  year={2025}
}

@article{chen2025xbench,
  title={xbench: Tracking Agents Productivity Scaling with Profession-Aligned Real-World Evaluations},
  author={Chen, Kaiyuan and Ren, Yixin and Liu, Yang and Hu, Xiaobo and Tian, Haotong and Xie, Tianbao and Liu, Fangfu and Zhang, Haoye and Liu, Hongzhang and Gong, Yuan and others},
  journal={arXiv preprint arXiv:2506.13651},
  year={2025}
}

@article{zhang2025rlvmr,
  title={Rlvmr: Reinforcement learning with verifiable meta-reasoning rewards for robust long-horizon agents},
  author={Zhang, Zijing and Chen, Ziyang and Li, Mingxiao and Tu, Zhaopeng and Li, Xiaolong},
  journal={arXiv preprint arXiv:2507.22844},
  year={2025}
}

@article{rahman2025spark,
  title={SPARK: Stepwise Process-Aware Rewards for Reference-Free Reinforcement Learning},
  author={Rahman, Salman and Gorantla, Sruthi and Gupta, Arpit and Roy, Swastik and Peng, Nanyun and Liu, Yang},
  journal={arXiv preprint arXiv:2512.03244},
  year={2025}
}

@article{xi2025agentprm,
  title={AgentPRM: Process Reward Models for LLM Agents via Step-Wise Promise and Progress},
  author={Xi, Zhiheng and Liao, Chenyang and Li, Guanyu and Yang, Yajie and Chen, Wenxiang and Zhang, Zhihao and Wang, Binghai and Jin, Senjie and Zhou, Yuhao and Guan, Jian and others},
  journal={arXiv preprint arXiv:2511.08325},
  year={2025}
}

@article{xu2025hybrid,
  title={Hybrid Reward Normalization for Process-supervised Non-verifiable Agentic Tasks},
  author={Xu, Peiran and Li, Zhuohao and Xing, Xiaoying and Zhang, Guannan and Li, Debiao and Shi, Kunyu},
  journal={arXiv preprint arXiv:2509.25598},
  year={2025}
}

@article{li2025one,
  title={One Model to Critique Them All: Rewarding Agentic Tool-Use via Efficient Reasoning},
  author={Li, Renhao and Tu, Jianhong and Su, Yang and Alinejad-Rokny, Hamid and Wong, Derek F and Lin, Junyang and Yang, Min},
  journal={arXiv preprint arXiv:2510.26167},
  year={2025}
}

@article{hu2025openreward,
  title={OpenReward: Learning to Reward Long-form Agentic Tasks via Reinforcement Learning},
  author={Hu, Ziyou and Shi, Zhengliang and Zhu, Minghang and Li, Haitao and Sun, Teng and Ren, Pengjie and Verberne, Suzan and Ren, Zhaochun},
  journal={arXiv preprint arXiv:2510.24636},
  year={2025}
}

@article{zhang2025linking,
  title={Linking Process to Outcome: Conditional Reward Modeling for LLM Reasoning},
  author={Zhang, Zheng and Shan, Ziwei and Song, Kaitao and Li, Yexin and Ren, Kan},
  journal={arXiv preprint arXiv:2509.26578},
  year={2025}
}

@article{jian2025patarm,
  title={PaTaRM: Bridging Pairwise and Pointwise Signals via Preference-Aware Task-Adaptive Reward Modeling},
  author={Jian, Ai and Ruan, Jingqing and Ma, Xing and Li, Dailin and Zhou, QianLin and Zeng, Ke and Cai, Xunliang},
  journal={arXiv preprint arXiv:2510.24235},
  year={2025}
}

@article{zhang2025bradley,
  title={Bradley-terry and multi-objective reward modeling are complementary},
  author={Zhang, Zhiwei and Liu, Hui and Li, Xiaomin and Dai, Zhenwei and Zeng, Jingying and Wang, Fali and Lin, Minhua and Chandradevan, Ramraj and Li, Zhen and Luo, Chen and others},
  journal={arXiv preprint arXiv:2507.07375},
  year={2025}
}

@article{zhang2025r1,
  title={R1-reward: Training multimodal reward model through stable reinforcement learning},
  author={Zhang, Yi-Fan and Lu, Xingyu and Hu, Xiao and Fu, Chaoyou and Wen, Bin and Zhang, Tianke and Liu, Changyi and Jiang, Kaiyu and Chen, Kaibing and Tang, Kaiyu and others},
  journal={arXiv preprint arXiv:2505.02835},
  year={2025}
}

@article{chen2025rm,
  title={Rm-r1: Reward modeling as reasoning},
  author={Chen, Xiusi and Li, Gaotang and Wang, Ziqi and Jin, Bowen and Qian, Cheng and Wang, Yu and Wang, Hongru and Zhang, Yu and Zhang, Denghui and Zhang, Tong and others},
  journal={arXiv preprint arXiv:2505.02387},
  year={2025}
}

@article{yang2025qwen3,
  title={Qwen3 technical report},
  author={Yang, An and Li, Anfeng and Yang, Baosong and Zhang, Beichen and Hui, Binyuan and Zheng, Bo and Yu, Bowen and Gao, Chang and Huang, Chengen and Lv, Chenxu and others},
  journal={arXiv preprint arXiv:2505.09388},
  year={2025}
}

@article{li2025webthinker,
  title={Webthinker: Empowering large reasoning models with deep research capability},
  author={Li, Xiaoxi and Jin, Jiajie and Dong, Guanting and Qian, Hongjin and Wu, Yongkang and Wen, Ji-Rong and Zhu, Yutao and Dou, Zhicheng},
  journal={arXiv preprint arXiv:2504.21776},
  year={2025}
}

@article{lu2025agentrewardbench,
  title={Agentrewardbench: Evaluating automatic evaluations of web agent trajectories},
  author={L{\`u}, Xing Han and Kazemnejad, Amirhossein and Meade, Nicholas and Patel, Arkil and Shin, Dongchan and Zambrano, Alejandra and Sta{\'n}czak, Karolina and Shaw, Peter and Pal, Christopher J and Reddy, Siva},
  journal={arXiv preprint arXiv:2504.08942},
  year={2025}
}

@article{shao2024deepseekmath,
  title={Deepseekmath: Pushing the limits of mathematical reasoning in open language models},
  author={Shao, Zhihong and Wang, Peiyi and Zhu, Qihao and Xu, Runxin and Song, Junxiao and Bi, Xiao and Zhang, Haowei and Zhang, Mingchuan and Li, YK and Wu, Yang and others},
  journal={arXiv preprint arXiv:2402.03300},
  year={2024}
}

@article{xia2025agent0,
  title={Agent0: Unleashing self-evolving agents from zero data via tool-integrated reasoning},
  author={Xia, Peng and Zeng, Kaide and Liu, Jiaqi and Qin, Can and Wu, Fang and Zhou, Yiyang and Xiong, Caiming and Yao, Huaxiu},
  journal={arXiv preprint arXiv:2511.16043},
  year={2025}
}

@article{lin2025comprehensive,
  title={A Comprehensive Survey on Reinforcement Learning-based Agentic Search: Foundations, Roles, Optimizations, Evaluations, and Applications},
  author={Lin, Minhua and Wu, Zongyu and Xu, Zhichao and Liu, Hui and Tang, Xianfeng and He, Qi and Aggarwal, Charu and Zhang, Xiang and Wang, Suhang},
  journal={arXiv preprint arXiv:2510.16724},
  year={2025}
}

@article{zhang2025critique,
  title={Critique-grpo: Advancing llm reasoning with natural language and numerical feedback},
  author={Zhang, Xiaoying and Sun, Hao and Zhang, Yipeng and Feng, Kaituo and Lu, Chaochao and Yang, Chao and Meng, Helen},
  journal={arXiv preprint arXiv:2506.03106},
  year={2025}
}

@article{feng2025video,
  title={Video-r1: Reinforcing video reasoning in mllms},
  author={Feng, Kaituo and Gong, Kaixiong and Li, Bohao and Guo, Zonghao and Wang, Yibing and Peng, Tianshuo and Wu, Junfei and Zhang, Xiaoying and Wang, Benyou and Yue, Xiangyu},
  journal={arXiv preprint arXiv:2503.21776},
  year={2025}
}

@article{li2025editthinker,
  title={EditThinker: Unlocking Iterative Reasoning for Any Image Editor},
  author={Li, Hongyu and Zhang, Manyuan and Zheng, Dian and Guo, Ziyu and Jia, Yimeng and Feng, Kaituo and Yu, Hao and Liu, Yexin and Feng, Yan and Pei, Peng and others},
  journal={arXiv preprint arXiv:2512.05965},
  year={2025}
}

@article{wu2025reinforcing,
  title={Reinforcing spatial reasoning in vision-language models with interwoven thinking and visual drawing},
  author={Wu, Junfei and Guan, Jian and Feng, Kaituo and Liu, Qiang and Wu, Shu and Wang, Liang and Wu, Wei and Tan, Tieniu},
  journal={arXiv preprint arXiv:2506.09965},
  year={2025}
}

@article{wang2025adatooler,
  title={AdaTooler-V: Adaptive Tool-Use for Images and Videos},
  author={Wang, Chaoyang and Feng, Kaituo and Chen, Dongyang and Wang, Zhongyu and Li, Zhixun and Gao, Sicheng and Meng, Meng and Zhou, Xu and Zhang, Manyuan and Shang, Yuzhang and others},
  journal={arXiv preprint arXiv:2512.16918},
  year={2025}
}

@article{whitehouse2025j1,
  title={J1: Incentivizing thinking in llm-as-a-judge via reinforcement learning},
  author={Whitehouse, Chenxi and Wang, Tianlu and Yu, Ping and Li, Xian and Weston, Jason and Kulikov, Ilia and Saha, Swarnadeep},
  journal={arXiv preprint arXiv:2505.10320},
  year={2025}
}

@article{fan2025sophiavl,
  title={SophiaVL-R1: Reinforcing MLLMs Reasoning with Thinking Reward},
  author={Fan, Kaixuan and Feng, Kaituo and Lyu, Haoming and Zhou, Dongzhan and Yue, Xiangyu},
  journal={arXiv preprint arXiv:2505.17018},
  year={2025}
}

@article{wang2024secrets,
  title={Secrets of rlhf in large language models part ii: Reward modeling},
  author={Wang, Binghai and Zheng, Rui and Chen, Lu and Liu, Yan and Dou, Shihan and Huang, Caishuang and Shen, Wei and Jin, Senjie and Zhou, Enyu and Shi, Chenyu and others},
  journal={arXiv preprint arXiv:2401.06080},
  year={2024}
}

@article{feng2025onethinker,
  title={OneThinker: All-in-one Reasoning Model for Image and Video},
  author={Feng, Kaituo and Zhang, Manyuan and Li, Hongyu and Fan, Kaixuan and Chen, Shuang and Jiang, Yilei and Zheng, Dian and Sun, Peiwen and Zhang, Yiyuan and Sun, Haoze and others},
  journal={arXiv preprint arXiv:2512.03043},
  year={2025}
}

@article{guo2025deepseek,
  title={Deepseek-r1: Incentivizing reasoning capability in llms via reinforcement learning},
  author={Guo, Daya and Yang, Dejian and Zhang, Haowei and Song, Junxiao and Zhang, Ruoyu and Xu, Runxin and Zhu, Qihao and Ma, Shirong and Wang, Peiyi and Bi, Xiao and others},
  journal={arXiv preprint arXiv:2501.12948},
  year={2025}
}

@article{liu2025understanding,
  title={Understanding r1-zero-like training: A critical perspective},
  author={Liu, Zichen and Chen, Changyu and Li, Wenjun and Qi, Penghui and Pang, Tianyu and Du, Chao and Lee, Wee Sun and Lin, Min},
  journal={arXiv preprint arXiv:2503.20783},
  year={2025}
}

@inproceedings{
anonymous2025mcpr,
title={{MCP}-R1: Generalized Real-World Task Agent Mastering Dozens of Tools},
author={Anonymous},
booktitle={Submitted to The Fourteenth International Conference on Learning Representations},
year={2025},
url={https://openreview.net/forum?id=1cQChGqNX4},
note={under review}
}

@article{tang2025rethinking,
  title={Rethinking Sample Polarity in Reinforcement Learning with Verifiable Rewards},
  author={Tang, Xinyu and Zhan, Yuliang and Li, Zhixun and Zhao, Wayne Xin and Zhang, Zhenduo and Wen, Zujie and Zhang, Zhiqiang and Zhou, Jun},
  journal={arXiv preprint arXiv:2512.21625},
  year={2025}
}

\clearpage
\appendix

\section{Dataset Details}

All datasets will be publicly released to support future research.

\subsection{Agent Training Data Distribution}

We collect 709k question-answer pairs from publicly available datasets as our RL dataset, \textbf{Reagent-RL-709K}. The detailed distribution information is shown in Figure \ref{fig:dataset} (bottom). We randomly select 100k data from RL dataset and utilize DeepSeekV3.1 to collect problem solving trajectories with our 6 tools. The trajectories reach to the correct final answer is saved as the SFT dataset in Figure \ref{fig:dataset} (top). In total, we collect 55.6k high quality trajectories for SFT training, denoted as \textbf{Reagent-SFT-55.6K}.

\subsection{Dataset Selection and Filtering}
\label{appendix:data selection}

\paragraph{DeepMath.}
Each question in DeepMath is accompanied by three independently generated solutions.
We remove samples for which the final answers are inconsistent across the three solutions, as such cases introduce ambiguity in supervision.

\paragraph{DeepScaleR.}
We filter out samples whose provided solutions produce answers inconsistent with the labeled ground-truth answer.

\paragraph{SimpleRL-Zoo.}
To encourage non-trivial reasoning behavior, we subsample questions by difficulty, retaining a higher proportion of medium- and hard-level questions and fewer easy ones.

\paragraph{MMK12.}
We select samples where the visual input consists of charts or tables, which can be reliably processed using OCR-based tool assistance.

\paragraph{PixelReasoner.}
We select questions that require extracting information from textual content or visual avatars within images.
These samples are solvable using a combination of OCR and image description tools.

\paragraph{LiveVQA.}
We retain questions that ask about identifiable attributes such as titles or authors present in images.
Only samples from the image-based subset are included.

\paragraph{ToolVQA.}
We select questions from the ImageDescription, GoogleSearch, OCR, and Calculator categories to align with the agent's available tool set.

\paragraph{SimpleDeepSearcher.}
We convert the original tool-calling format into a Qwen-compatible format to ensure consistency with the agent’s action space.

\paragraph{AFM-WebAgent.}
We transform multi-agent interaction data into a single-agent reasoning trajectory by linearizing planning, verification, and reflection steps.
These components provide useful reasoning patterns for single-agent reasoning.

\paragraph{LongAudio.}
We select audio samples with durations between 5 and 40 seconds.
Audio clips shorter than 5 seconds typically lack sufficient informational content, while longer clips impose excessive computational overhead on the audio-to-text model (whisper-large-v3).

\subsection{\rewardmodel{} Construction Details}
\label{appendix:reward model data}

The prompt template employed to generate training data for \rewardmodel{} is detailed in Figure~\ref{fig:prompt_template}. Our primary design objective is to augment the reasoning-driven analytical capabilities of \rewardmodel{}, ensuring it provides reliable and informative feedback at both the semantic and scalar levels. To this end, each training instance is structured into three distinct components to facilitate multi-granular reasoning supervision:
\begin{itemize}[topsep=2pt, itemsep=2pt, parsep=0pt]
    \item \texttt{<think>}: A reasoning process that evaluates the logical consistency of the given trajectory. By explicitly articulating the rationale behind the assessment, this component provides transparency into how the reward model derives its judgments, thereby enhancing the interpretability of the final reward signal.
    \item \texttt{<critique>}: A targeted summary that identifies reasoning flaws, with a particular focus on global logic and the appropriate invocation of external tools.
    \item \texttt{<score>}: A holistic scalar value derived from the preceding analysis to quantify the trajectory's overall quality. By condensing complex reasoning evaluations into a standardized numerical format, this component serves as the formal reward signal required for advantage calculation within the reinforcement learning optimization loop.
\end{itemize}

To support the development of our model, we curate two specialized datasets: \textbf{Reagent-RRM-SFT-28K}, comprising 28,000 high-quality trajectories for initial supervised fine-tuning, and \textbf{Reagent-RRM-RL-90K}, consisting of 90,000 instances designed for large-scale RL training.

\begin{figure*}[h]
\begin{promptbox}{Instruction Prompt for Reward Model Annotation}
\setlength{\hsize}{\textwidth}
\small
You are an expert agent tool use evaluator. You must strictly follow the output format below:\\

\texttt{<think>} \\
Provide a comprehensive analysis of the entire reasoning trajectory. Focus specifically on the agent’s reasoning quality and its tool-usage behavior across ANY type of tool.

Key points to evaluate (for all tasks and all tools):\\
- Whether the agent correctly decided when to call tools. Over-reliance on tools for trivial reasoning is bad; failing to call tools when necessary is also bad.\\
- Whether the agent misused tools (e.g., calling an irrelevant tool, giving incorrectly formatted arguments, hallucinating tool inputs or filenames, making repeated tool calls without new purpose).\\
- Whether the agent understood tool limitations (e.g., tool outputs may be incomplete, noisy, or partial; tools cannot access nonexistent resources).\\
- Whether the agent improved its reasoning over time (e.g., corrected wrong assumptions, avoided repeated mistakes, verified hypotheses when possible).\\
- Whether the agent avoided unverified guesses. Hypotheses without verification are harmful.\\
- Whether the agent avoided fabricating tool results, file names, object identifiers, or other non-existent content.\\

If uncertain, identify potential harmful reasoning patterns: unnecessary tool calls, missing essential tool calls, uncritical acceptance of tool output, faulty logical jumps, or incorrect assumptions about tool capabilities.\\
Never mention the true answer. Only evaluate the reasoning process and tool use.\\
\texttt{</think>} \\

\texttt{<critique>} \\
Provide a succinct, specific, and actionable summary of issues in the agent’s reasoning and tool use. This section will be shown to the agent, so it must be concise and clearly highlight:

- Incorrect, unnecessary, missing, or repeated tool calls.\\
- Incorrect assumptions, unverified reasoning, or blind trust in tool results.\\
- Any improper handling of tool limitations or constraints.\\
- Any hallucinated tool arguments, filenames, or resource identifiers.\\
- Unlogical reasoning.\\
Do NOT provide the correct answer or hints toward it.\\
\texttt{</critique>} \\

\texttt{<score>} \\
A single float between 0 and 1 representing the overall quality of the reasoning and tool use.
0 means completely incorrect or harmful reasoning; 1 means flawless reasoning with appropriate, precise, and well-justified tool use.
\texttt{</score>} \\

\textbf{Strict Requirements:} Output exactly three blocks; focus solely on reasoning/tool-use; never reveal the correct answer.
\end{promptbox}
\caption{The prompt used for generating structured judgments of reward model.}
\label{fig:prompt_template}
\end{figure*}

\section{Training Details}
\label{appendix:training}

\subsection{Training Codebase}

We use LLaMA-Factory~\cite{zheng2024llamafactory} to implement SFT training of both \rewardmodel{} and Reagent. Both reward model and agent model are trained for 2 epoches.

We use rLLM~\cite{rllm2025} to implement Agentic RL training of Reagent. We use VeRL~\cite{sheng2024hybridflow} to implement RL training of \rewardmodel{}. The hyper-parameters we used during RL training is shown in Table~\ref{tab:hyperparams}.
We conduct RL training for 300 steps.

\begin{table}[htbp]
\centering
\small
\caption{Hyper-parameters for Reinforcement Learning Training.}
\label{tab:hyperparams}
\begin{tabular}{llc}
\toprule
\textbf{Category} & \textbf{Hyper-parameter} & \textbf{Value} \\
\midrule
\multirow{5}{*}{Training Config} 
& Base Model & Qwen3-8B \\
& Optimizer & AdamW \\
& Learning Rate & $5 \times 10^{-7}$ \\
& Training Batch Size & 64 \\
& Mini-batch Size & 16 \\
\midrule
\multirow{6}{*}{Generation \& Env} 
& Max Agent Steps & 13 \\
& Temperature & 0.7 \\
& Top-$p$ & 0.95 \\
& Rollout Samples ($n$) & 8 \\
& Lambda ($\lambda$) & 0.3 \\
\bottomrule
\end{tabular}
\end{table}

\subsection{Tools configuration}
\label{appendix:tool train}
Our models are trained on 8 NVIDIA A800-80G GPUs.
We detail the specific implementations of tools integrated into our agentic framework during training:

\begin{itemize}[leftmargin=*]
\item \textbf{Search}: Powered by the Bing Search API. The agent receives the top-$k$ results, including the URL, title, and a content snippet for each entry.
\item \textbf{Browse}: Website content is retrieved via the Jina Reader and subsequently condensed using DeepSeek-Chat as a summarization model.
\item \textbf{Image2text}: Visual queries and image-based reasoning are handled by GPT-4.1.
\item \textbf{Audio2text}: Audio inputs are transcribed into text using the Whisper-large-v3 model.
\end{itemize}

\section{Evaluation Details}
\label{appendix:eval}

\subsection{Evaluation Benchmarks}

For GAIA, we report performance on the 103-task \texttt{Text} subset in Table~\ref{tab:comprehensive_performance} following \citet{dong2025agentic}, while the full set results (165 tasks) are reported as GAIA (Full). For HLE, we evaluate on the 500-task subset consistent with \citet{dong2025agentic}, as shown in Table~\ref{tab:comprehensive_performance}. For xbench, we report results specifically on the \texttt{xbench-DeepSearch} set.

\subsection{Tool Configurations}

The configurations for evaluation tools are identical to those detailed in Appendix~\ref{appendix:tool train}.

\subsection{Agent Evaluation Settings}

For all benchmark evaluations, we employ a decoding temperature of 0.6 and a $top\_p$ of 0.95. The maximum response length is constrained to 32,768 tokens, and the agent is permitted a maximum of 30 tool-use steps per query. Consistent with \citet{dong2025agentic}, we utilize Qwen2.5-72B-Instruct as the automated judge to perform binary scoring by comparing the agent's predictions against the ground-truth answers.

\section{Case Study}
\label{appendix:case study}

\begin{figure*}
    \centering
    \includegraphics[width=\linewidth]{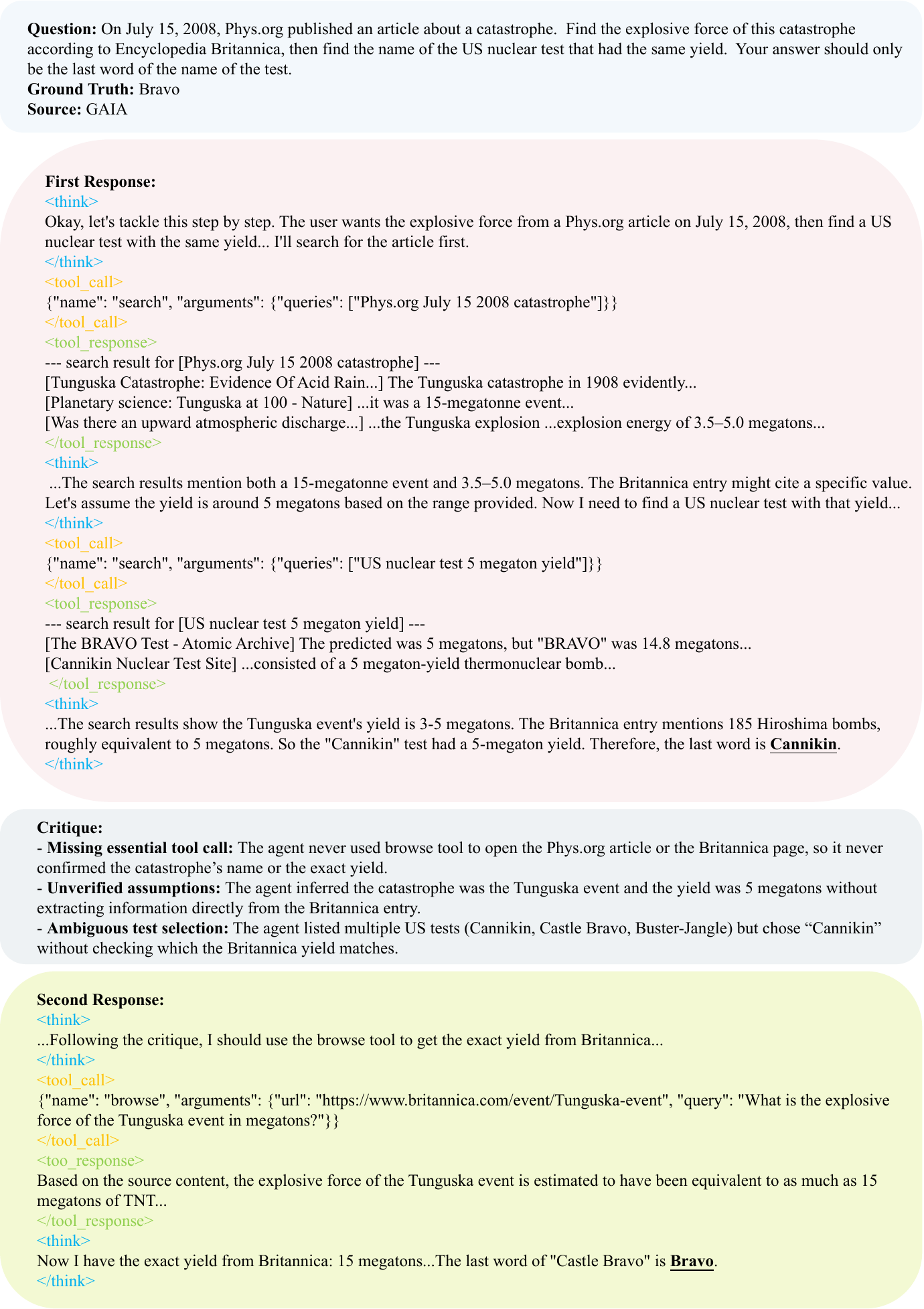}
    \caption{Case 1: a search question from GAIA.}
    \label{fig:case1}
\end{figure*}

\begin{figure*}
    \centering
    \includegraphics[width=\linewidth]{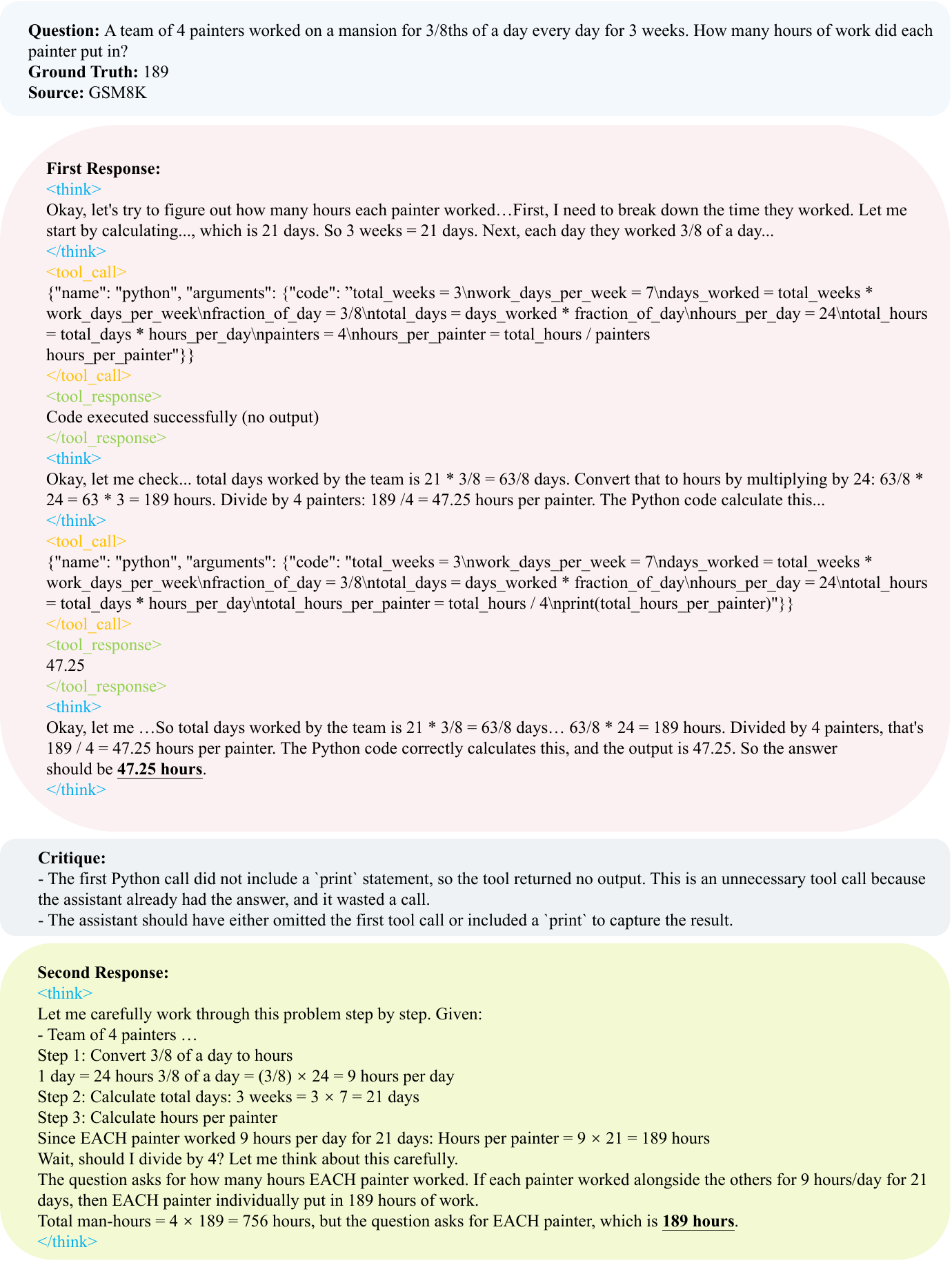}
    \caption{Case 2: a math question from GSM8K.}
    \label{fig:case2}
\end{figure*}

We present two case studies of Reagent-C in Figure~\ref{fig:case1} and Figure~\ref{fig:case2} to demonstrate the effectiveness of our critique mechanism. The critiques generated by Agent-RRM highlight logical inconsistencies or inappropriate tool usage in the initial responses. These critiques serve as informative signals that facilitate the agent's self-correction, leading to successful reasoning refinement and error rectification in the subsequent responses.

\end{document}